\documentclass[lettersize,journal]{IEEEtran}
\usepackage{amsmath,amsfonts}
\usepackage{algorithmic}
\usepackage{algorithm}
\usepackage{array}
\usepackage[caption=false,font=normalsize,labelfont=sf,textfont=sf]{subfig}
\usepackage{textcomp}
\usepackage{stfloats}
\usepackage{url}
\usepackage{verbatim}
\usepackage{graphicx}
\usepackage{cite}
\usepackage{multirow}
\begin{document}

\title{Latent Code Augmentation Based on Stable Diffusion for Data-free Substitute Attacks}

\author{Mingwen~Shao,~\IEEEmembership{Member,~IEEE,} 
	Lingzhuang~Meng, 
	Yuanjian~Qiao, 
	Lixu~Zhang,
	and~Wangmeng~Zuo,~\IEEEmembership{Senior Member,~IEEE} 
        % <-this % stops a space
\thanks{This work has been submitted to the IEEE for possible publication. Copyright may be transferred without notice, after which this version may no longer be accessible.}
\thanks{Corresponding authors: Lingzhuang Meng.}
\thanks{
	Mingwen Shao, Lingzhuang Meng, Yuanjian Qiao and Lixu Zhang are with the School of Computer Science and Technology, China University of Petroleum (East China), Qingdao, 266580, China (e-mail: smw278@126.com; lzhmeng1688@163.com; yjqiao@s.upc.edu.cn; z2216842477@163.com;)}
\thanks{Wangmeng Zuo is with the Harbin Institute of Technology, China (e-mail: wmzuo@hit.edu.cn;)}
\thanks{Manuscript received ***, 2024; revised ***, 2024.}}

% The paper headers
\markboth{Journal of \LaTeX\ Class Files,~Vol.~14, No.~8, *~2024}%
{Shao \MakeLowercase{\textit{et al.}}: Latent Code Augmentation for Data-free Substitute Attacks based on Diffusion Model}

\IEEEpubid{0000--0000/00\$00.00~\copyright~2024 IEEE}
% Remember, if you use this you must call \IEEEpubidadjcol in the second
% column for its text to clear the IEEEpubid mark.

\maketitle

\begin{abstract}
Since the training data of the target model is not available in the black-box substitute attack, most recent schemes utilize GANs to generate data for training the substitute model. However, these GANs-based schemes suffer from low training efficiency as the generator needs to be retrained for each target model during the substitute training process, as well as low generation quality.
To overcome these limitations, we consider utilizing the diffusion model to generate data, and propose a novel data-free substitute attack scheme based on the Stable Diffusion (SD) to improve the efficiency and accuracy of substitute training.
Despite the data generated by the SD exhibiting high quality, it presents a different distribution of domains and a large variation of positive and negative samples for the target model.
For this problem, we propose Latent Code Augmentation (LCA) to facilitate SD in generating data that aligns with the data distribution of the target model. Specifically, we augment the latent codes of the inferred member data with LCA and use them as guidance for SD. 
With the guidance of LCA, the data generated by the SD not only meets the discriminative criteria of the target model but also exhibits high diversity. 
By utilizing this data, it is possible to train the substitute model that closely resembles the target model more efficiently.
Extensive experiments demonstrate that our LCA achieves higher attack success rates and requires fewer query budgets compared to GANs-based schemes for different target models. 
Our codes are available at \url{https://github.com/LzhMeng/LCA}.
\end{abstract}


\section{Introduction}
\IEEEPARstart{C}{onvolutional} Neural Networks (CNNs) currently exhibit remarkable performance and play a pivotal role in a wide range of applications in various fields. However, recent studies have shown that CNNs are vulnerable to adversarial perturbations, which may lead to incorrect decisions and pose risks in critical applications~\cite{yuan_adversarial_2019,MIM,UAP}. For example, a little imperceptible adversarial perturbations can cause autonomous vehicles to make wrong decisions, leading to severe consequences. Such adversarial perturbation reveal the vulnerability of CNNs. Therefore, there is currently an increasing focus on adversarial perturbations, adversarial attacks, and their defence~\cite{katzir_gradients_2021,AdverDefen}.

\begin{figure*}[!t]
	\centering
	\includegraphics[width=\linewidth]{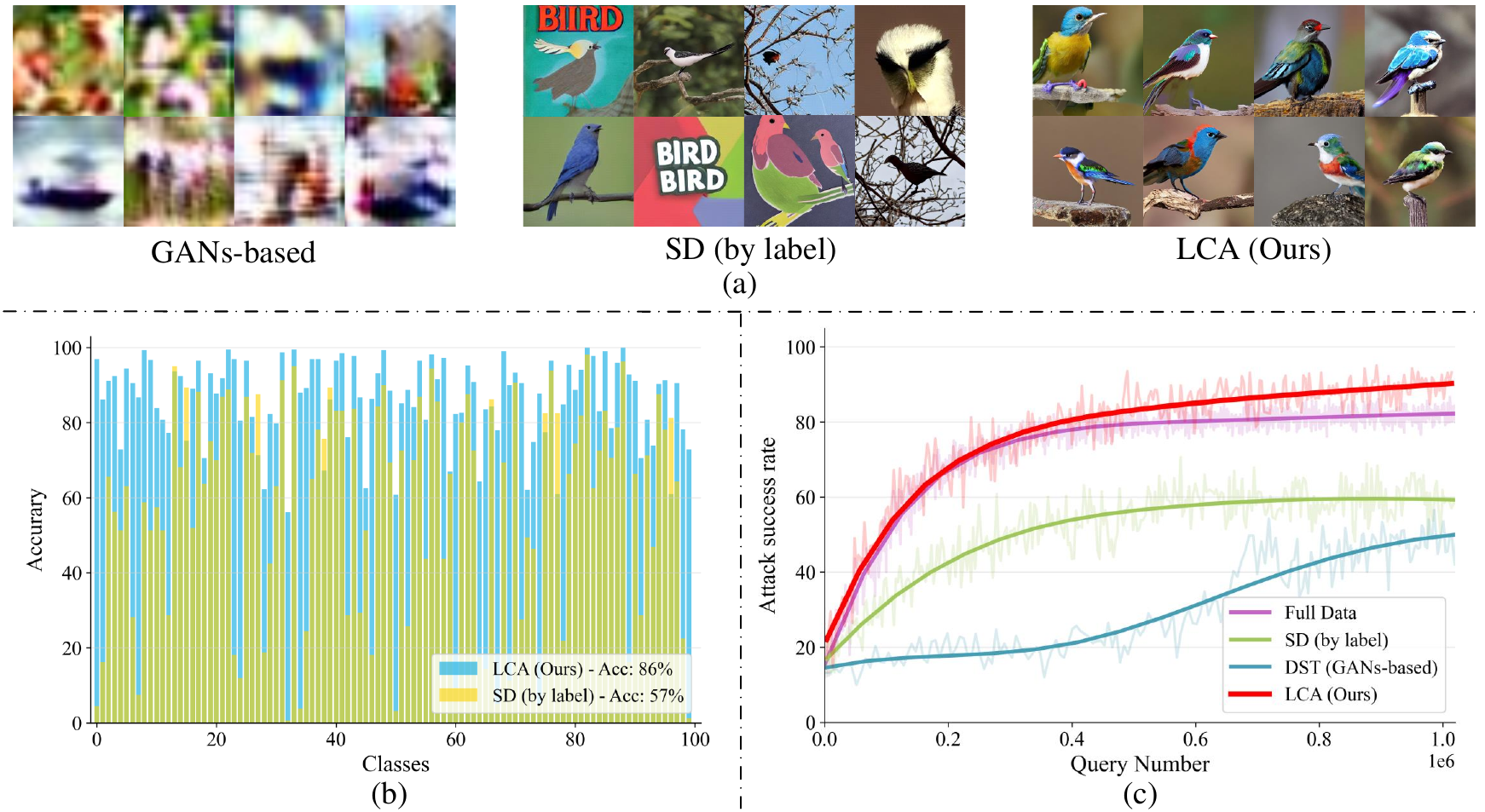}
	\caption{Comparison of our scheme with others. \textbf{(a) Visualisation of the generated data.} Includes data generated by GANs-based schemes during substitute training, Stable Diffusion prompted by class labels, and our LCA. \textbf{(b) The accuracy of each class and average accuracy.} The accuracy of the data generated directly with the Stable Diffusion varies widely between classes, with some classes being less than 10\%, whereas the data generated by our LCA is more homogeneous and has a higher accuracy rate. \textbf{(c) Attack success rate.} Our LCA achieves a higher attack success rate with fewer queries. Where full data indicates substitute training with the full training set data of the target model.}
	\label{fig_1}
\end{figure*}
Existing adversarial attack schemes can be categorized into white-box attacks and black-box attacks according to the availability of network information. The white-box attacks assume complete knowledge of the network structure and parameters, allowing for the effective generation of adversarial samples using gradient-based methods, such as FGSM~\cite{FGSM}, BIM~\cite{BIM}, PGD~\cite{PGD}, etc. However, in realistic scenarios where networks are deployed on servers, there is no access to information about the network, making black-box attacks more practical~\cite{yue2021black,EBA,TSEA,MAZE,DaST,DST,HardLabel,ColorQuery,queries,9727149}.
The black-box attacks aim to achieve attacks on the target networks using only the output (label~\cite{HardLabel} or probability~\cite{MAZE}) of the target networks, and researchers have proposed various solutions for this purpose. For example, transfer-based attacks~\cite{EBA,TSEA} exploit the transferability of adversarial samples, enabling adversarial samples generated on other models to be applied to the black-box target networks. Another example is the query-based attacks~\cite{ColorQuery,queries,9727149}, which adds small noise to the image each time and keeps querying the target model output until the target result is obtained. 
In addition, substitute-based attack~\cite{MAZE,DaST,DST,TED} requires training the substitute network similar to the target network and then using the adversarial samples generated by the substitute network to attack the black-box target network. 
In this paper, we focus on the substitute-based approach which often achieve better attack success rates than the other schemes. The essential reason is that the substitute model can be very similar to the target model through effective training, making the attack on the substitute network very close to the direct attack on the white-box target model~\cite{DST}. Therefore, training the substitute model that is close to the target model is a key issue for the substitute-based approach.

\IEEEpubidadjcol

The substitute-based schemes utilize knowledge distillation methods to make the output of the substitute model fit the output of the target model~\cite{MAZE,DaST}.
In this substitute training process, if the original training data of the target model is available, it is possible to train the substitute network that closely resembles the target network. %By leveraging the substitute model, adversarial samples with higher attack success rates can be generated against the target model.
%This is because deep learning models are primarily influenced by data, as indicated by their data-driven nature~\cite{}.
However, in the realistic scenario, we do not have access to the original training data of the target model.
In light of this, existing solutions invariably use data generated by Generative Adversarial Networks (GANs) to train the substitute models~\cite{MAZE,DaST,DDG,DST,TED}. For example, Zhou et al.~\cite{DaST} proposed DaST to solve the sample imbalance problem using a generator with a multi-branch structure, and Wang et al.~\cite{DDG} encouraged GANs to learn the data distribution of different classes.
However, these GANs-based methods suffer from low training efficiency, as the generator needs to be retrained for each target model and the quality of the generated images is relatively low.
In recent years, Diffusion Models (DMs)~\cite{DDPM,StableDiffusion} have demonstrated superior generation quality as well as diverse generation results, which can directly generate images of various categories without requiring additional targeting training as in GANs. These advantages motivate us to consider \textit{utilizing pre-trained Stable Diffusion (SD)}~\cite{StableDiffusion} \textit{to generate richer data for training the substitute model.}

%Nevertheless, when directly using the diffusion model for substitute training, we observed that \textbf{the generated data exhibits various domain distributions and contains many samples that do not conform to the discriminative criteria of the target model}, as shown in Fig.~\ref{fig_1} (a). 
Despite SD having demonstrate excellent performance in image generation tasks, we observe that the diversity of the generated data is too rich, which led to two problems when used directly for substitute training. \textbf{On the one hand, the generated images cover different domain distributions and most of the samples do not match the domain distribution of the target model.} As shown in Fig.~\ref{fig_1} (a), the images generated using the SD prompted by class labels cover a variety of types, such as drawings, landscapes, close-ups, and text. This resulted in the generation of a large number of negative samples that do not contribute positively to substitute training, bringing about a futile increase in resources.
\textbf{On the other hand, there is a significant disparity in the number of positive and negative samples between different classes in the generated data}. As depicted in Fig.~\ref{fig_1} (b), the accuracy of some classes is less than 10\% when evaluated in the target network.  This class imbalance problem in the data affects the performance of the substitute training.
%Using such imbalanced data for substitute training can result in a notable performance gap between the substitute model and the target model, especially for classes with particularly small positive samples.

To alleviate the above problems, we propose a novel Latent Code Augmentation (LCA) method to facilitate the SD to generate data that conforms to the data distribution of the target model.
Specifically, we first generate some data with the help of SD guided by text prompts with class names. Subsequently, we introduce Membership Inference (MI) to identify the data that are most likely to belong to the training data of the target model, known as member data. 
The latent codes of the member data are augmented by our LCA and serve as guidance for the SD.
Thanks to the LCA guidance, the SD is able to generate images that are consistent with the data distribution of the member data.
Experimental results demonstrate that our LCA is able to significantly improve the substitute training efficiency and outperforms the existing state-of-the-art (SOTA) substitute attack solutions based on GANs in scenarios where no training data from the target model is available, as illustrated in Fig.~\ref{fig_1}(c). Moreover, it even outperforms the case with training data, due to the sufficient abundance of generated data.

The main contributions of this paper are as follows:
\begin{itemize}	
	\item To the best of our knowledge, we are the first to leverage the SD to improve the performance of data-free substitute attacks, making substitute training more efficient and achieving higher attack success rates.
	\item We propose LCA to augment the latent code and use it as guidance for the SD to further facilitate the generation of data that better matches the data distribution of the target network.
	\item The experimental results demonstrate that our LCA outperforms GANs-based schemes in terms of attack success rates and query efficiency across different target networks trained on various training sets.
\end{itemize}

\section{Related Work}
\label{sec:Rela}  
\subsection{Adversarial Attacks} 
Adversarial attacks can be categorized into white-box and black-box attacks according to whether the structure and parameters of the target network are available. For white-box attacks~\cite{FGSM,BIM,PGD,DistrWhite}, the network gradient or data distribution is mainly utilized to generate adversarial samples and usually achieves a high success rate of attacks. Classical examples include FGSM~\cite{FGSM}, BIM~\cite{BIM}, PGD~\cite{PGD}, which take a gradient perspective and add perturbations along the gradient direction of the target category to generate adversarial samples. In addition, Zhu et al.~\cite{DistrWhite} reconceptualized the adversarial sample in terms of data distribution, by manipulating the distribution of the images to induce misclassifications. However, these approaches rely on model parameter information, limiting their application in real-world scenarios.

For black-box attacks~\cite{chen2017zoo,hu2022substitute,zhou2022adversarial}, the adversarial samples are generated by utilizing only the limited output (labels~\cite{HardLabel} or probabilities~\cite{MAZE}) of the target model and are therefore more realistic. One solution to black-box attacks is transfer-based attacks~\cite{EBA,TSEA,duan_dual_2023}, which enhance the transferability of adversarial samples generated using white-box attacks on known networks to attack the black-box target model. For example, Li et al.~\cite{Sibling} integrated multiple different models to craft adversarial samples in order to achieve better transferability.
Another solution is substitute-based attacks~\cite{MAZE,DDG,DaST,DST,cui2020substitute,park2020partial}, which focus on training the substitute model using the output of the target model. Then the adversarial samples generated using the white-box attacks on the substitute model can then be used to attack the target network.
There are also other solutions, such as query-based attacks~\cite{ColorQuery,queries,zhu2022defense,9727149}, which add small perturbations to the image at a time and then query the output of the target network until the target network makes an incorrect judgment. Among black-box attack schemes, substitute attacks have gained significant attention because they achieve good performance in both target and non-target attacks when the substitute model closely mimics the target model.

\subsection{Data-free Substitute Attacks} 
In real-world scenarios, it is usually difficult to obtain the training data of CNNs deployed on servers, which makes data-free substitute attacks more difficult to implement~\cite{yue2021black}. Existing research~\cite{MAZE,DDG,DaST,DST} has explored generating data with GANs and then using these data to train the substitute network that simulates the target network.

Among them, in order to generate samples with a more uniform distribution of classes, Zhou et al.~\cite{DaST} designed label-controlled generators for each class so that the generated samples cover all classes of the attacked model more evenly. In contrast, Wang et al.~\cite{DDG} embedded labels in the layers of the generator so that the generator learns relatively independent data distributions for each class to synthesize diverse images.
To better adapt to different target models, Wang et al.~\cite{DST} implemented a dynamic substitute network using gate structures and used graph structures to optimize the distillation process.
To better optimize the generator, Kariyappa et al.~\cite{MAZE} used a zero-gradient estimation technique to estimate the gradient of the target model so that the generator can be optimized indirectly in the black-box setting. In addition, Zhang et al.~\cite{TED} changed the game between the generator and the substitute model by optimizing the generator and the substitute model independently instead of jointly for more stable convergence to the target model.

The aforementioned schemes invariably rely on GANs as the main structure to generate the data used for training the substitute model. However, the generator needs to be re-optimized for each target model during substitute training, resulting in poor quality of the generated images and inefficient training. In order to improve the efficiency of substitute training, we utilize the pre-trained DMs to generate data, which has the advantage of generating diverse and high-quality data from different domains without re-training.

\subsection{Diffusion Model} 
DMs~\cite{DDPM,StableDiffusion} are trained on large-scale datasets and have gained attention for their ability to generate realistic, natural, and diverse images. It is divided into two steps: forward diffusion and reverse denoising. In the diffusion process, the clean image $x$ is added with $T$ steps of noise $\epsilon$ to obtain the noisy image $x_{t}$. In the denoising process, the denoising encoder $\epsilon^{_{\theta }}$ in the form of a U-Net is used to estimate the noise added at step $t$. During the sampling process, the diffusion model can infer the noise of the given noisy image using the denoising encoder $\epsilon^{_{\theta }}$, which eventually generates clean and natural images through $T$-step denoising.

The objective function of the network at step $t$ can be expressed as:
\begin{equation}
	L_{DM}=\mathbb{E}_{x,\epsilon\sim \mathcal{N}(0,1),t}\left [ \left \| \epsilon-\epsilon^{_{\theta }}\left ( x_{t},t \right ) \right \|_{2}^{2} \right ].
\end{equation}

To improve the efficiency of the diffusion model, Rombach et al.~\cite{StableDiffusion} extended the diffusion model from the image space to the latent space $z$, called Stable Diffusion. It can generate images with specific requirements conditioned on either text or images, where text is encoded by the pre-trained CLIP text encoder and images are encoded by the pre-trained `AutoEncoderKL' image encoder. The encoding of the condition $\tau _{\theta}(y)$ is integrated into the U-Net architecture through the cross-attention. The objective function of SD can be expressed as:
\begin{equation}
	L_{LDM}={\mathbb{E}}_{z,y,\epsilon\sim \mathcal{N}(0,1),t}\left [ \left \| \epsilon-\epsilon^{_{\theta }}\left ( z_{t},t,\tau _{\theta}(y) \right ) \right \|_{2}^{2} \right ].
\end{equation}

\subsection{Membership Inference} 
MI is an attack scheme in the privacy leakage category that determines whether a given sample belongs to the training data set of the model, i.e. the member data~\cite{LOGAN,GANLeaks}. MI exploits the observation that CNNs are constantly fitting the training data during training and respond differently to the training and non-training data. Therefore, using the overfitting property of the model, MI can infer membership using the output of the model. 

In the MI scheme for attacking image classification, Salem et al.~\cite{MLLeaks} utilized the prediction confidence of the model as a credential to infer membership. They argue that higher prediction confidence indicates a higher possibility of belonging to the training set. However, this approach relies on the confidence output of the model and is not suitable for black-box models that only provide label outputs.
To overcome this limitation, Choquette-Choo et al.~\cite{LabelonlyMI} presented the label-only MI scheme that inferred sample membership based on whether the model made the correct decision when faced with a slight perturbation (e.g., adversarial perturbations). The theoretical basis of the scheme lies in the fact that the CNNs continuously fit the training data during the training phase so that the training data are contained within the decision boundary as much as possible, while the non-training data are more likely to be close to the boundary. That is, the training data has higher robustness than the non-training data, and therefore data with higher robustness is more likely to belong to the member data.

In this paper, we introduce MI to identify data that is most likely to belong to the training data of the target model and use this member data as the reference for DMs to generate data more suitable for the target model.

\begin{figure*}[!t]
	\centering
	\includegraphics[width=\linewidth]{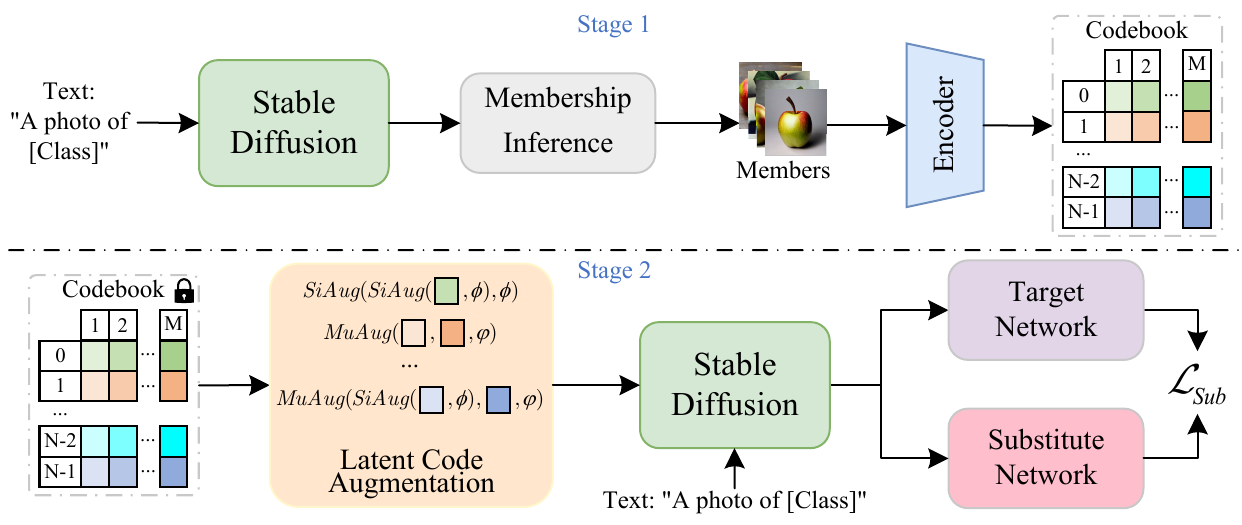}
	\caption{The framework of our scheme, which is divided into two stages. \textbf{Stage 1:} Inferring member data that matches the distribution of the target model and encoding it into the codebook. \textbf{Stage 2:} Guiding SD to generate good images and training the substitute models. Where `Encoder' is the image encoder `AutoEncoderKL' in the pre-trained SD. The number of classes in the codebook is N and the length is M. $SiAug$ and $MuAug$ are single-code augmentation and multi-code augmentation functions, respectively. $\phi$ and $\psi$ are different augmentation operations. $\mathcal{L}_{Sub}$ is the loss of substitute training.}
	\label{fig_2}
\end{figure*}
\section{Method}    
In this section, we provide a comprehensive description of our data-free substitute attack scheme based on the SD guided by latent code augmentation. First, the problem formalization is given in Sec.~\ref{Formalization} and the overview of the complete scheme is presented in Sec.~\ref{Overview}. Subsequently, we describe in detail the process of inferring membership data and our LCA approach in  Sec.~\ref{IMD} and Sec.~\ref{LCA}. Finally, the loss function of substitute training is described in Sec.~\ref{ST}.

\subsection{Problem Formalization}\label{Formalization}
In this paper, we follow the substitute attack paradigm to implement black-box attacks. Specifically, given a black-box target model $T$, we train a substitute model $S$ to be as similar as possible to $T$. 
Subsequently, adversarial samples are generated by applying white-box adversarial sample generation methods to the substitute model and then used to attack the target model. The process of training substitute model is known as substitute training and can be formalized as:
\begin{equation}
    \arg\mathop{\min}\limits_{S} \mathbb{E}_{x\sim\mathcal{X}} \mathcal{D}(T(x),S(x)),
\end{equation}
where $\mathcal{D}$ denotes the difference measure between the target and the substitute models, and $\mathcal{X}$ denotes the training dataset of the target model.

Subsequently, the adversarial samples are generated from the substitute models using white adversarial sample generation methods. We use PGD~\cite{PGD} as an example:
\begin{equation}
	x_{adv} = clip(x_{adv} + \alpha \cdot sign(\nabla_{x_{adv}} L), \epsilon),
\end{equation}
where $\nabla_{x_{adv}} L$ is the gradient of the loss function $L$ of the adversarial sample $x_{adv}$ against the ground truth, $\alpha$ is the step parameter, $clip(\cdot,\epsilon)$ is the clipping function restricting the perturbation to the $\epsilon$ range.
These generated adversarial samples are used to attack the target model.

\subsection{Overview}\label{Overview}
Since the training data of the target model is not accessible, we propose a novel SD-based scheme for substitute training to generate data that matches the target model distribution. We divide the process of our scheme into two stages: 1) inferring member data that matches the distribution of the target model and encoding it into the codebook, and 2) guiding the SD to generate data and training the substitute model, the framework is shown in Fig.~\ref{fig_2}. 

In Stage 1, we first generate some data using the pre-trained SD guided by the text prompts $Text$ of the class name:
\begin{equation}
	\mathbf{X}=SD(Text).
\end{equation}

%We simplify all diffusion, denoising steps, and mapping to image space operations to $SD$ to represent the Stable Diffusion generating images based on prompts of $Text$.
Subsequently, We utilize MI to infer these data and identify the samples that are most likely to belong to the member data of the target model, as described in Sec.~\ref{IMD}. These data are encoded and stored into the codebook, denoted by $codebook=\{c_{n,1},c_{n,2},\cdots,c_{n,M}\},n\in\{0,1,\cdots,N-1\}$, where $M$ is the codebook size and $N$ is the number of classes. The process of encoding the member data $\mathbf{X}_{memb}$ using the encoder $\varepsilon$ in SD can be written as:
\begin{equation}
	c_n=\varepsilon(\mathbf{X}_{memb}).
\end{equation}

After obtained the codebook, in Stage 2, we augment the latent codes leveraging our LCA and utilize it to guide the SD to generate data. The specific method is introduced in Sec.~\ref{LCA}, which can be described as:
\begin{gather}
	z_{aug}=LCA(codebook),\\
	\mathbf{X}_{aug}=SD(z_{aug},Text).
\end{gather}

%Guided by latent code conditions, the data generated by the SD retains the features of the member data at the feature level and exhibits good diversity. 
Finally, the continuously generated data $\mathbf{X}_{aug}$ is utilized to train the substitute model. The process of substitute training with this data is shown in Sec.~\ref{ST}.

\subsection{Inferring Member Data}\label{IMD}
In order to distinguish which data matches the data distribution of the target model, we introduce MI to infer the member data from the data generated by SD. We directly add a small amount of noise to the samples, and distinguish between member and non-member data through the network's fluctuation for the noisy images. 

Specifically, we first add a small amount of Gaussian noise to the clean images $x$ to obtain the noisy image $\hat{x}$:
\begin{equation}
	\hat{x}=x+\mathcal{N}(0,\sigma ^{2}),
\end{equation}
where, $\mathcal{N}(0,\sigma ^{2})$ denotes Gaussian noise with a mean of $0$ and a variance of $\sigma ^{2}$.
Subsequently, we introduce the decision distance, which represents the degree of membership $dist_T$ in terms of the network's decision distance between the clean and noise samples:
\begin{equation}
	\label{eqmi}
	dist_T (x)=\mathcal{D}(T(x),T(\hat{x})),
\end{equation}
where $\mathcal{D}$ is the decision distance of the two samples. Specifically, for scenarios with probabilistic outputs, we employ Mean Squared Error (MSE) to measure the distance between the two sample outputs, and samples with small distances are more likely to belong to the training data. The member data can be obtained by:
\begin{equation}
	\label{memb}
    \mathbf{X}_{memb}=\{x|dist_T (x)\le u\},
\end{equation}
where $u$ is a threshold. For scenarios with only label outputs, we classify the samples with unchanged outputs after adding noise as member data, while those with changed outputs are considered as non-member samples. This process can be described as:
\begin{equation}
	\label{memb2}
	\mathbf{X}_{memb}=\{x|T(x)=T(\hat{x})\}.
\end{equation}

The effectiveness of Eq. (\ref{eqmi}) lies in the fact that the network exhibits different responses to small perturbations among various membership data~\cite{LabelonlyMI}. Moreover, during training, the network tends to include training data within the decision center, while non-training data is closer to the decision boundary. Therefore, if the network's response to a very small perturbation in the given data is drastic or even results in misclassification, then that data is more likely to be non-training data.
%One of the benefits of using the robustness of the network to infer member data is that the inferred data is highly robust and able to avoid data bias during latent code augmentation.

\begin{figure*}[!t]
	\centering
	\includegraphics[width=\linewidth]{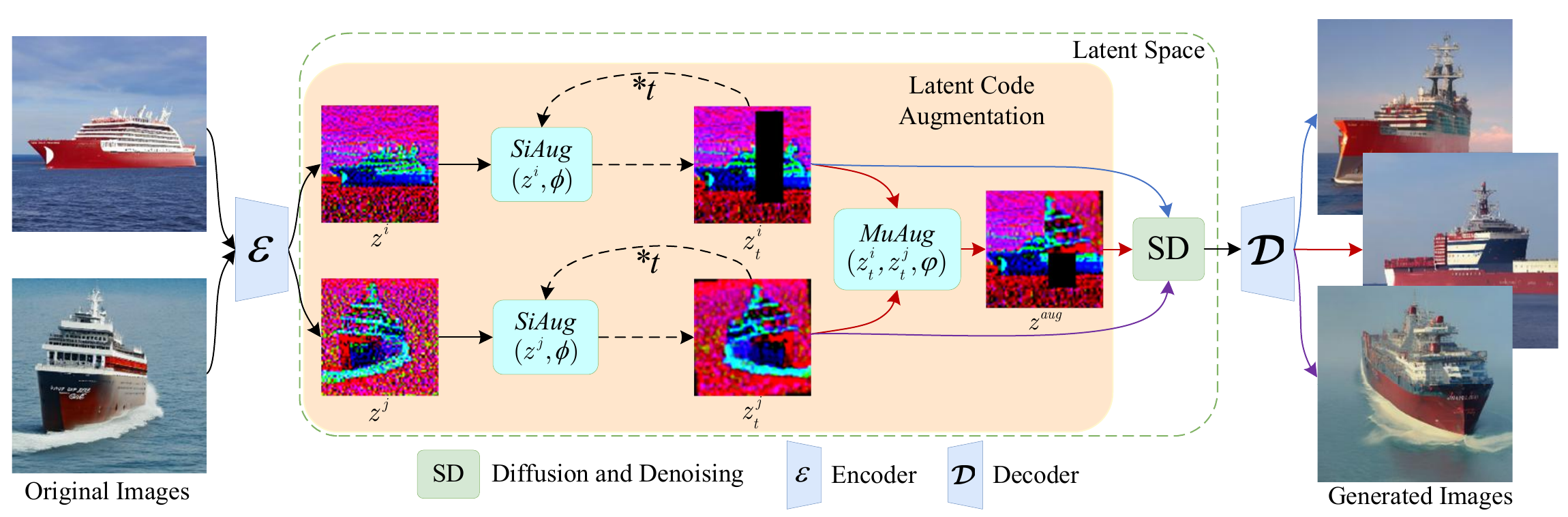}
	\caption{The process of generating data using Latent Code Augmentation. Our LCA augments the latent code of the image, which is then used to guide the SD in generating data. Where $SiAug$ and $MuAug$ are single-code augmentation and multi-code augmentation functions, respectively, $t$ is the number of iterations, and $\phi$ and $\psi$ are different code augmentation operations.}
	\label{fig_3}
\end{figure*}
\subsection{Latent Code Augmentation}\label{LCA}
%To enhance the diversity of the generated images while maintaining the distinctive features of the member data, we introduce the latent code augmentation (LCA) approach.
In order to generate images that match the data distribution of the target model, we propose a novel LCA to guide the SD for generating the data, the process is shown in Fig.~\ref{fig_3}.
The image-to-image generation method in SD is to code a reference image into the latent space then generate an image guided by the latent code. On this basis, we encodes the inferred few member data, augments them in latent space, and then uses the augmented latent code as guidance for SD to generate data. Our LCA aims to guide SD in the latent space to generate data that maintain the characteristics of the member data with good diversity.
%By augmenting latent code, we introduce diversity into latent encoding. These augmented codes are then utilized to guide the diffusion model in the generation of diverse and high-quality images. 
%The LCA approach aims to explore ways to enable diffusion models in the latent space to generate more diverse outputs while retaining the essential features of the member data.

The augmentation in the latent space is similar to traditional data augmentation, we first define the set of single-code augmentation operations $\Phi = \{\phi_1, \phi_2, \ldots, \phi_k\}$, which includes Translation, Padding, Rotation, Cropping, Scaling, Affine, Erasing, Gaussian blurring, Gaussian noise, Pepper \& Salt noise, and so on. Then, we use this set to iteratively perform a series of augmentation  operations on the latent code:
\begin{equation}
	z_{t}=SiAug(z_{t-1},\phi_t), \phi_t \in \Phi,
\end{equation}
where $z$ denotes latent code in latent space, $t$ is the number of iterations for augmentation, we take $t\le k$, and $\phi_t$ denotes the operation in the $t$-th round of augmentation. After each iteration, the operation is removed from the set of single operations to ensure that each operation is performed once:
\begin{equation}
	\Phi = \Phi \verb|\| \{\phi_t \},
\end{equation}
where `$\verb|\|$' denotes the difference operation of sets, i.e. removing the element $\phi_t$ from the set $\Phi$.
In addition, each augmentation operation can employ a different magnitude of augmentation. The iterative single-image augmentation motivates us to be able to generate diverse latent codes.

Then, we define the set of multi-code augmentation operations $\Psi=\{\psi_1,\psi_2,\dots,\psi_l\}$ that include MixUp~\cite{MixUp}, CutMix~\cite{CutMix}, RICAP~\cite{RICAP}, etc. And the latent codes $z^i$ and $z^j$ are further fused based on $\Psi$ to further expand the diversity of the latent codes:
\begin{equation}
	z^{aug}=MuAug(z^i,z^j,\psi),\psi \in \Psi.
\end{equation}

By fusing the latent codes that have been augmented with single-code augmentation as well as randomly selecting single-code augmentation and multi-code augmentation, a variety of different latent codes can be obtained.
The algorithm is described in Algorithm~\ref{alg:alg1}.
\renewcommand{\algorithmicrequire}{\textbf{Inputs:}}
\renewcommand{\algorithmicensure}{\textbf{Output:}}
\begin{algorithm}[t]
	\caption{Steps of latent code augmentation.}\label{alg:alg1}
	\begin{algorithmic}[1]
		\REQUIRE $codebook$: the codebook with original latent codes.%; $\Phi$: the set of single-code augmentation operations; $\Psi$: the set of multi-code augmentation operations,
		\ENSURE $z_{aug}$: the augmented latent code.
		\STATE Initialize the set of single-code augmentation operations $\Phi=\{\phi_1,\phi_2,\dots,\phi_k\}$ and the set of multi-code augmentation operations $\Psi=\{\psi_1,\psi_2,\dots,\psi_l\}$;
		\STATE Randomly select single- or multi-code augmentation;
		\IF{single-code augmentation}
		\STATE Randomly sample latent codes $z_0$ in codebook;
		\STATE Sample iteration number $t$ with $t<k$;
		\FOR{$i=1$ to $t$}
		\STATE Random sampling $\phi_i \in \Phi$;
		\STATE $z_{i}=SiAug(z_{i-1},\phi_i)$;
		\STATE $\Phi=\Phi \verb|\| \{\phi_i \}$;
		\ENDFOR
		\STATE $z^{aug}$ = $z_t$;
		\ELSIF{multi-code augmentation}
		\STATE Randomly sample $z_0^i,z_0^j$ in $codebook$ for the same class and augment to obtain $z_t^i,z_t^j$ with reference to single-code augmentation (steps 4 to 10);
		\STATE Random sampling $\psi \in \Psi$;
		\STATE $z^{aug}=MuAug(z_t^i,z_t^j,\psi)$;
		\ENDIF
		\STATE \textbf{return}  $z_{aug}$
	\end{algorithmic}
\end{algorithm}
%Guided by these augmented latent codes, the SD generates images with higher diversity while preserving important features of the member data such as the data distribution.
%This allows for the creation of diverse combinations of latent codes, resulting in a broader range of visually diverse outputs.Furthermore, by randomly selecting and applying various augmentations from set , the generated latent codes exhibit higher diversity.  Guided by these latent codes, the diffusion model generates images with enhanced diversity while preserving the significant features of the member data. 

\textbf{But is the augmented latent code still a meaningful and valid latent code?}

As we know, the convolution layer $f$ maps images to feature vectors, and the translational weight tying in CNNs gives it translation equivalence~\cite{equiv4,equiv}. That is, a translation $\phi$ of the image $I$ leads to corresponding translation $\pi$ of the feature map $f(I)$:
\begin{equation}
	f(\phi(I))=\pi(f(I)).
\end{equation}

In general, $\pi \neq \phi$, due to the presence of pooling layers.
However, similar to the generator in VAE, `AutoEncoderKL' does not have the pooling layer and therefore has equivalents such as rotation equivalence and translation equivalence~\cite{equiv1}. In the case of ignoring edge effects, the feature vector of the transformed image is equivalent to the transformed feature vector of the image by a specific scale, i.e. $\pi=\phi$.

This implies that augmenting the latent code is equivalent to encoding the augmented image. As a result, the augmented latent codes remain meaningful and valid latent codes of image. These augmented codes can be used as guidance of the SD to generate diverse and visually appealing images while preserving the distinctive attributes of the member data.

\begin{figure}[!t]
	\centering
	\includegraphics[width=\linewidth]{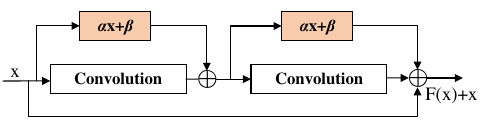}
	\caption{ResBlock with learnable parameters in the substitute model. $\alpha$ and $\beta$ are learnable parameters that are used to make the substitute model applicable to different target models.}
	\label{fig_3_5}
\end{figure}
\subsection{Substitute Training}\label{ST}
The data continuously generated by the LCA-guided SD in Stage 2 is used to train the substitute model. In order to adapt the substitute model to different target models. In this paper, we design the substitute model with ResNet34~\cite{ResNet} structure and introduce learnable parameters in each ResBlock to increase the adaptivity. The structure of the ResBlock with learnable parameters is shown in Fig.~\ref{fig_3_5}. 

To make the output of the substitute model $S$ similar to that of the target model $T$, we exploit both the cross-entropy loss and the MSE loss between the two networks:
\begin{align}
	\label{loss}
	\mathcal{L}_{Sub}=\mathbb{E}_{x\sim \mathbf{X}_{Aug}}&[ \lambda_{1}CE(S(x),T(x))\notag\\
	&+\lambda_{2}MSE(S(x),T(x))],
\end{align}
where $\lambda_{1}$ and $\lambda_{2}$ are hyperparameters. For the case  where only label outputs of the target model are available, we set $\lambda_{2} = 0$. By minimizing the loss, the substitute model is optimized to be more similar to the target model. The algorithm of overall scheme is described in Algorithm~\ref{alg:alg2}.
\renewcommand{\algorithmicrequire}{\textbf{Inputs:}}
\renewcommand{\algorithmicensure}{\textbf{Output:}}
\begin{algorithm}[t]
	\caption{Training steps of the proposed scheme.}\label{alg:alg2}
	\begin{algorithmic}[1]
		\REQUIRE $T$: the target model; $SD$: the Stable Diffusion; $M$: the lenght of the codebook; $K$: the max epoch.
		\ENSURE $S$: the substitute model.
		\STATE Initialize the $codebook=\{c_{n,1},c_{n,2},\cdots,c_{n,M}\}$,
		\STATE // Stage 1
		\FOR{$n$=0 to $N-1$}
		\STATE $\mathbf{X} = SD(Text)$ //Generates data using SD;
		\STATE $\mathbf{X}_{memb}=\{x|dist_T (x)\le u\}$ //Infer the member data in each classes using Eq.~(\ref{memb} or \ref{memb2});
		\STATE $c_n=\varepsilon(\mathbf{X}_{memb})$ //Encode the member data and update it into coderbook.
		\ENDFOR
		\STATE // Stage 2
		\FOR{$k$ =1 to $K$}
		\STATE $z_{aug}=LCA(codebook)$ //Obtain augmented codes according to Alg. \ref{alg:alg1};
		\STATE $\mathbf{X}_{Aug} = SD(z^{aug},Text)$ //Generate data based on the SD with two guiding conditions;
		\STATE Input $\mathbf{X}_{Aug}$ into the target and substitute networks and optimize $S$ based on the loss $\mathcal{L}_{Sub}$ in Eq. (\ref{loss}).
		\ENDFOR
		\STATE \textbf{return}  $S$
	\end{algorithmic}
\end{algorithm}

\section{Experiments}
\label{experiment}
In this section we provide the implementation details of the experiment (Sec.~\ref{ED}), the resulting image generated by latent code augmentation (Sec.~\ref{LCAR}), the performance in different cases of black-box substitute attacks (Sec.~\ref{BAR}), and the ablation experiments with different components and settings (Sec.~\ref{AS}). 

\subsection{Experiment Details}\label{ED}
\subsubsection{Datasets and models} We evaluated the performance of our LCA on different target networks, include VGG (16, 19)~\cite{VGG}, ResNet (18, 34, 50)~\cite{ResNet}. The training datasets for the target models include CIFAR10~\cite{CIFAR} (10 classes, image size 32$\times$32), CIFAR100~\cite{CIFAR} (100 classes, image size 32$\times$32), STL10~\cite{STL10} (10 classes, image size 96$\times$96) and Tiny-ImageNet~\cite{Tinyimagenet} (200 classes, image size 64$\times$64). 

In this paper, we use the pre-trained Stable Diffusion~\cite{StableDiffusion} as the primary generator, which is a latent text-to-image diffusion model trained on $512\times 512$ images from a subset of the LAION database~\cite{LAION-5B}. This model utilizes the frozen CLIP ViT-L/14~\cite{CLIP} text encoder to guide the model based on text prompts. We use the v1 version of SD in both stages, which generate high-quality diversity images at $512\times 512$ resolution. The difference between the two stages is the content of the guidance, where the first stage utilizes text prompts consisting of class labels as guidance, and the second stage employs latent codes and text prompts as dual guidance. Since the SD is not trained or fine-tuned on the training data of the target model in this paper, our LCA remains a data-free substitute attack scheme.

\subsubsection{Training setup} All training and testing is performed on GPU: NVIDIA GeForce RTX 2080 Ti. The threshold $u$ in Eq.~(\ref{memb}) is set to $1e-3$. The substitute network is optimized by $Adam$ with the initial learning rate of 0.001 and batch size set to 32.
For a fair comparison, we set the query budget, i.e., the number of queries to the target model, uniformly in the comparison schemes: 500k queries on the CIFAR10 and STL10 datasets, 1M queries on the CIFAR100 dataset, and 3M queries on the Tiny-ImageNet datasets.

\subsubsection{Evaluation process} For a target network, we generate adversarial samples by using gradient-based white-box adversarial sample generation methods, FGSM~\cite{FGSM}, PGD~\cite{PGD} and BIM~\cite{BIM}, on the substitute models trained in this paper. The Attack Success Rate (ASR) of these adversarial samples against the target model is used as the final evaluation index. The ASR can be calculated by the following formula:
\begin{equation}
	ASR=\frac{N_{suc}}{N_{all}},
\end{equation}
where $N_{suc}$ is the number of successful attack adversarial samples and $N_{all}$ is the total number of generated adversarial samples.

\begin{figure*}[!t]
	\centering
	\includegraphics[width=\linewidth]{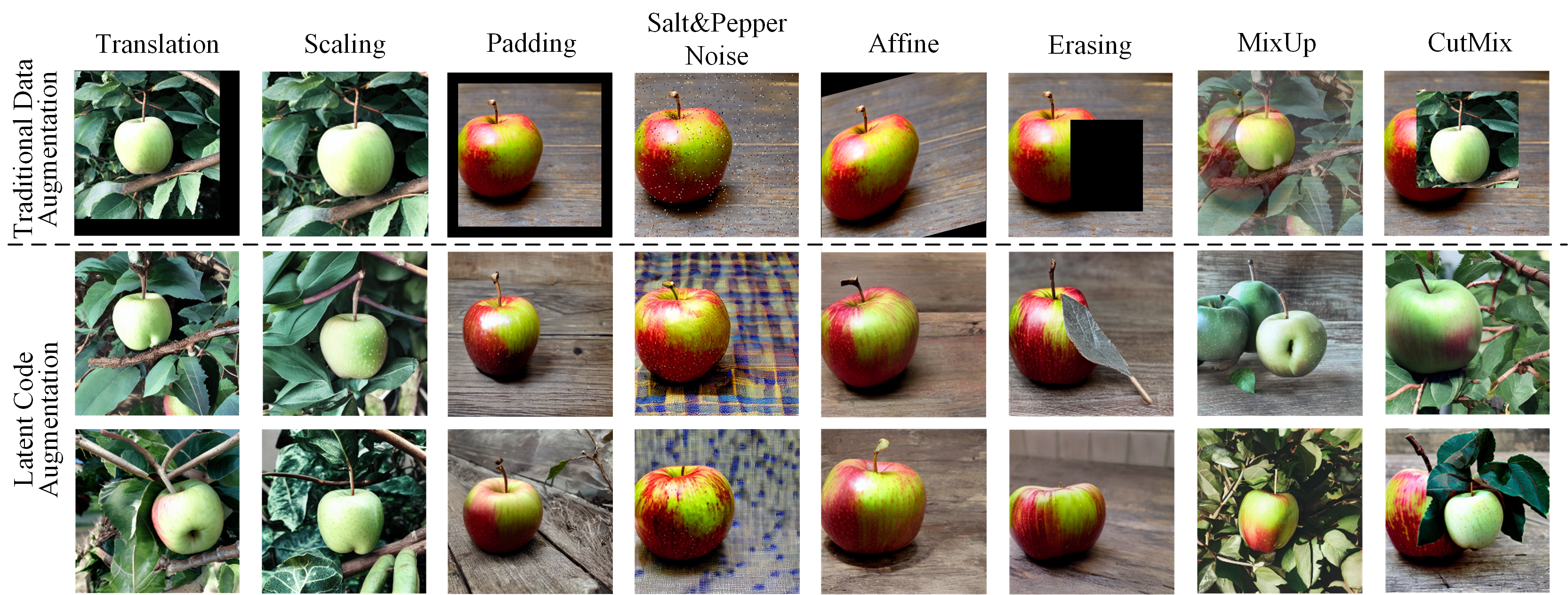}
	\caption{Visual comparison of latent code augmentation and traditional data augmentation. Only a single augmentation operation is performed. The image generated by LCA retains the features of the original image and is more natural than traditional data augmentation.}
	\label{fig_4}
\end{figure*}
\begin{figure*}[!t]
	\centering
	\includegraphics[width=\linewidth]{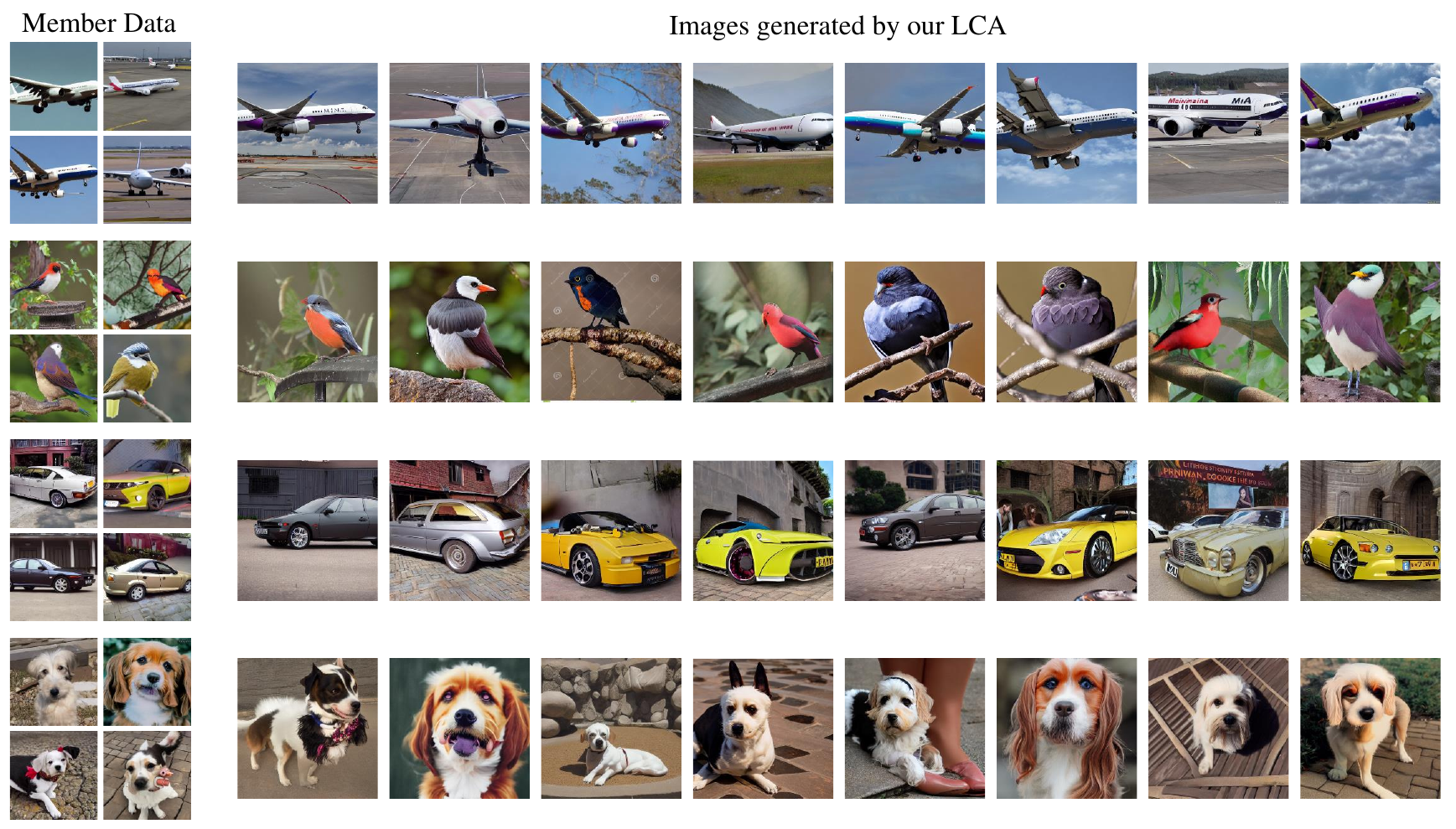}
	\caption{The images generated by our LCA. The generated data has a wider diversity along with high features consistency with member data.}
	\label{fig_5}
\end{figure*}
\subsection{Visualization results of LCA}\label{LCAR}
\subsubsection{Visual effects of latent code augmentation} 
We present the results after augmentation of latent codes, as shown in Fig.~\ref{fig_4}. We utilize a series of manipulations to augment the latent code, which in turn generate diverse images using the SD. 
It can be seen that the images obtained by operations such as Translation, Padding, and Affine transformation do not exhibit black edges compared to the traditional data augmentation, which indicates that the image effect obtained by our LCA is smooth and seamless.
In addition, the images generated by adding noise to the latent codes are altered in texture while maintaining the natural appearance of the foreground and background. This provides an effective means of introducing variation into the image while preserving the overall integrity of the image as compared to traditional data augmentation.
%Moreover, the images generated by erasing and CutMix operations show natural and visually coherent fillings.

Moreover, it can be observed that the effect of augmentation obtained by a single transformation applied to a single image is limited. This ensures that the essential characteristics of the original image are retained throughout the augmentation process.
In contrast, the image obtained by fusing two different scales of latent codes exhibits greater diversity. For example, the images obtained using MixUp and CutMix fuse latent codes of different scales to generate images with richer variation. At the same time, it displays a great advantage over traditional data augmentation in terms of visual quality and naturalness.

Furthermore, by applying random augmentation on a larger number of latent codes, the images generated by the SD exhibit higher diversity, as depicted in Fig.~\ref{fig_5}. It is important to emphasize that these generated images successfully retain features such as the distribution of member data while providing better diversity, which is superior to traditional data augmentation. This is attributed to the fact that 1) the SD at the latent space level generates more natural images, in addition, 2) the LCA augments the latent codes with a wider range of combinations, generating images with higher diversity.
Overall, our LCA provides a more efficient approach for task-specific image generation.

\subsubsection{Accuracy of latent code augmentation} 
We present the accuracy bar graphs for each data class obtained through LCA in Fig.~\ref{fig_1} (b). Which is the accuracy results obtained from images generated according to CIFAR100 and classified using the ResNet50 target network. The comparison is the data generated by the SD using class labels as the prompt, which have relatively low average accuracy and uneven distribution of positive and negative samples. In contrast, our LCA demonstrates higher average accuracy and a more uniform distribution of positive samples across classes. This reflects the necessity of our LCA, which generates data that are closer to the data distribution of the target network and have a more balanced number of positive samples.

\begin{table*}[!t]
	\centering
	\renewcommand\arraystretch{1.1}
	\caption{Comparison of the attack success rates of our method and competitors on different target models. For a fair comparison, we use the same query budget. The PGD is used as the default attack method. The numbers in `()' indicate the average $L_2$ perturbation distance. The best result are shown in \textbf{bold}. \label{table1}}
	\resizebox{\linewidth}{!}{
		\begin{tabular}{c|c|ccc|cc|cc|c}
			\hline
			& Dataset       & \multicolumn{3}{c|}{CIFAR10} & \multicolumn{2}{c|}{CIFAR100} & \multicolumn{2}{c|}{STL10}   & \multicolumn{1}{c}{Tiny-ImageNet}    \\ \cline{2-10} 
			& Target Model                        & VGG16                & ResNet18             & ResNet34             & VGG19                & ResNet50             & ResNet34                     &   ResNet50                    &   ResNet50   \\ \hline
			\multirow{4}{*}{\rotatebox{90}{Non-Target}} & DaST~\cite{DaST}    & 48.65(1.28)          & 49.40(1.08)          & 51.80(1.44)          & 30.24(4.74)          & 26.41(4.20)          & 54.41(3.55)          & 60.50(2.88)          & 27.08(4.70) \\
			& DST~\cite{DST}      & 52.21(1.27)          & 54.93(1.10)          & 62.80(1.44)          & 34.00(4.70)          & 31.63(4.57)          & 66.00(3.60)          & 69.65(2.90)           & 32.51(4.69)\\
			& MAZE-PD~\cite{MAZE} & 64.39(1.20)          & 78.99(1.11)          & 66.80(1.34)          & 40.75(4.55)          & 44.44(4.32)          & 69.86(3.35)          & 73.73(2.88)           & 45.46(4.71) \\
			& LCA (Ours)                           & \textbf{82.54(1.16)} & \textbf{90.29(0.91)} & \textbf{89.64(1.13)} & \textbf{95.15(4.18)} & \textbf{97.82(4.32)} & \textbf{96.16(3.15)} & \textbf{96.97(2.89)} & \textbf{84.05(4.66)}\\ \hline
			\multirow{4}{*}{\rotatebox{90}{Target}}     & DaST~\cite{DaST}    & 29.29(1.24)          & 41.60(1.17)          & 33.64(2.24)          & 11.55(4.80)          & 20.04(5.12)          & 45.33(4.05)          & 50.09(3.99)          & 19.88(4.88)  \\
			& DST~\cite{DST}      & 32.50(1.23)          & 42.15(1.15)          & 35.26(2.25)          & 17.22(4.98)          & 22.54(4.98)          & 56.00(3.99)          & 66.22(4.11)    & 22.47(4.87)       \\
			& MAZE-PD~\cite{MAZE} & 33.20(1.24)          & 41.99(1.08)          & 39.05(1.20)          & \textbf{22.75(5.21)}          & 24.65(5.10)          & 56.39(4.05)          & 67.18(3.92)         & 23.55(4.90)  \\
			& LCA (Ours)                           & \textbf{35.19(1.21)} & \textbf{48.36(0.99)} & \textbf{39.15(1.20)} & 19.42(3.02) & \textbf{51.63(3.19)} & \textbf{79.93(3.97)} & \textbf{80.76(3.92)} & \textbf{26.79(4.88)}\\ \hline
		\end{tabular}
	}
\end{table*}
\begin{figure}[!t]
	\centering
	\includegraphics[width=\linewidth]{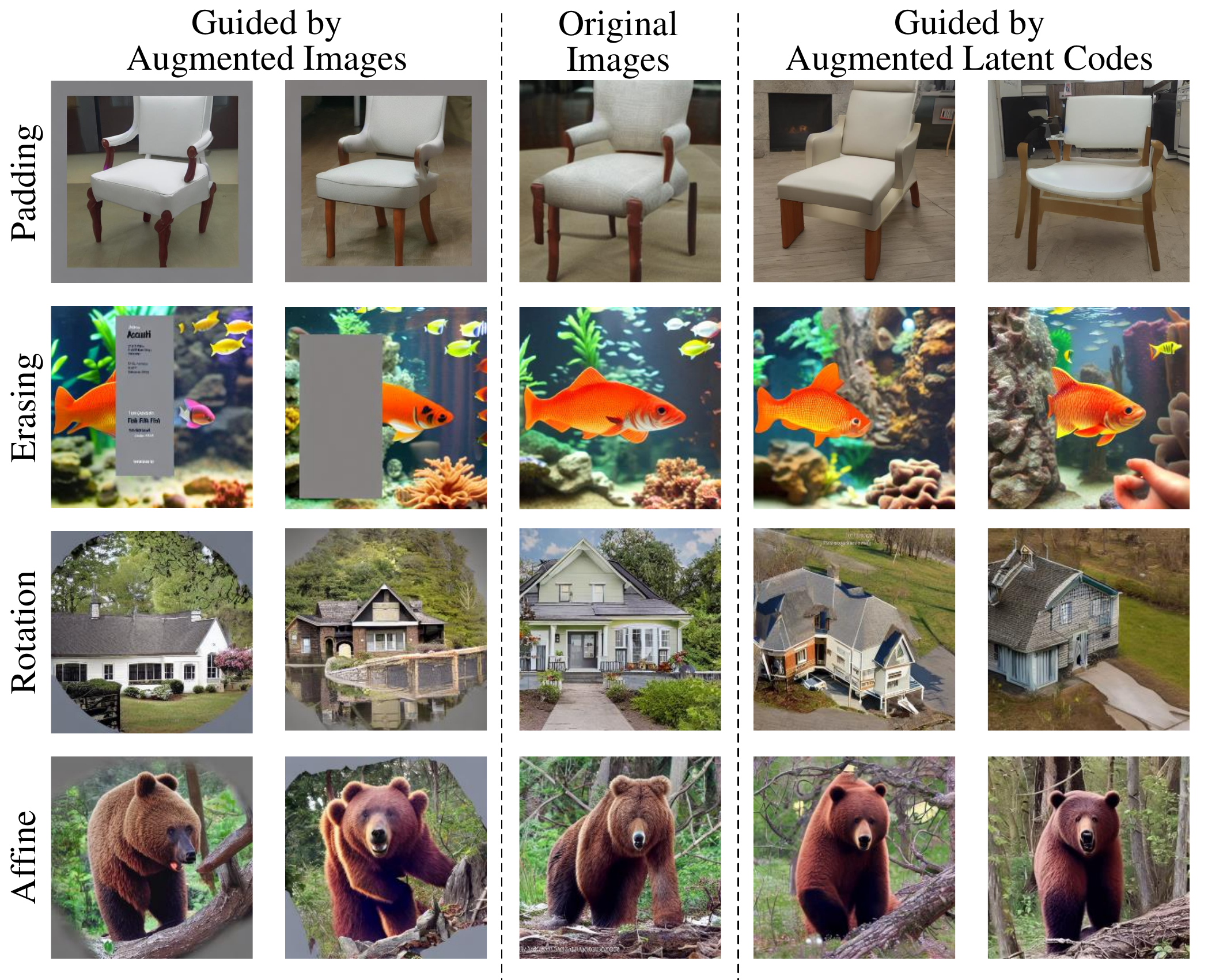}
	\caption{
		Comparison of SD-generated images guided by augmented images and augmented latent codes. Our LCA generates more natural images with no blank regions.}
	\label{fig_55}
\end{figure}
\subsubsection{Effectiveness of augmented latent code} 
In this section, we discussed why we chose to use augmented latent codes as the guidance of SD instead of augmented images. The images generated under two scenarios are presented in Fig.~\ref{fig_55}. It can be observed that using augmented images as guidance for SD leads to issues such as image blank regions and unnatural transitions, which are not present in our LCA.
This is because information such as blank regions after conventional image augmentation is encoded into the latent code. These blank regions are then considered as part of the image information by the SD, which generates the images containing these blank regions.
In contrast, in our LCA, the blank regions in the latent codes are regarded as missing features by SD. With the excellent internal drawing capability of the SD, the regions corresponding to the missing or distorted features are restored, resulting in highly natural images. Therefore, by applying LCA at the feature level, we avoid the problem of image blanking and ensure that the generated images do not exhibit unnatural artifacts or artificial traces.

\subsection{Attack Results of LCA}\label{BAR}
\subsubsection{ASR on different target models}
We report the ASR against various target models on different datasets in Table~\ref{table1}. The experiments include both target and non-target attack scenarios. It can be seen that our approach achieves higher ASR than other data-free baseline methods~\cite{DaST,DST} on all datasets while having lower or comparable $L_2$ distances. In addition, the performance of our LCA remains competitive compared to the partial data scheme MAZE-PD~\cite{MAZE}. Notably, our LCA achieves significant improvement in ASR for datasets with more classes such as CIFAR100, even at smaller $L_2$ distances. 
This is mainly attributed to the exceptional generative capability of SD, which generate high-quality and diverse images without the need for retraining. In contrast, GANs-based approaches~\cite{DaST,DST,MAZE} require fine-tuning of the generator during training, resulting in lower efficiency and accuracy under the same query budget. Therefore, in future research on data-free substitute attacks, DMs are undoubtedly a superior choice compared to GANs.
Furthermore, our LCA also plays a crucial role, as it generates data that is very similar to the original training data distribution of the target network. This further enhances the similarity between the trained substitute network and the target network and ultimately achieving superior performance.

\begin{table*}[!t]
	\centering
	\renewcommand\arraystretch{1.1}
	\caption{Comparison of the attack success rates of our approach and competitors on different white-box adversarial sample generation methods. For a fair comparison, we use the same query budget. We use ResNet18 as the target model for CIFAR10, ResNet50 as the target model for CIFAR100, and ResNet34 as the target model for STL10. The number in `()' indicates the average $L_2$ perturbation distance, with the best result being \textbf{bolded}. \label{table2}}
	\resizebox{\linewidth}{!}{
		\begin{tabular}{c|c|ccc|ccc|ccc}
			\hline
			& Dataset                             & \multicolumn{3}{c|}{CIFAR10}                                       & \multicolumn{3}{c|}{CIFAR100}                                      & \multicolumn{3}{c}{STL10}                                          \\ \cline{2-11} 
			& Target Model                        & FGSM                 & BIM                  & PGD                  & FGSM                 & BIM                  & PGD                  & FGSM                 & BIM                  & PGD                  \\ \hline
			\multirow{4}{*}{\rotatebox{90}{Non-Target}} & DaST~\cite{DaST}    & 39.12(1.54)          & 60.45(0.99)          & 49.40(1.08)          & 24.73(4.34)          & 31.10(4.34)          & 26.41(4.20)          & 40.42(3.90)          & 89.77(3.09)          & 54.41(3.55)          \\
			& DST~\cite{DST}      & 45.39(1.54)          & 78.54(0.98)          & 54.93(1.10)          & 30.42(4.53)          & 33.96(4.08)          & 31.63(4.57)          & 56.90(3.60)          & 96.84(3.14)          & 66.00(3.60)           \\
			& MAZE-PD~\cite{MAZE} & 66.57(1.53)          &90.97(0.98)          & 78.99(1.11)          & 34.07(4.34)          & 36.45(4.05)          & 44.44(4.32)          & 68.78(3.69)          & 97.68(3.41)          & 69.86(3.35)          \\
			& LCA (Ours)                          & \textbf{87.62(1.42)} & \textbf{93.33(0.94)} & \textbf{90.29(0.91)} & \textbf{95.14(4.23)} & \textbf{97.59(4.06)} & \textbf{97.82(4.32)} & \textbf{96.16(3.40)} & \textbf{98.18(3.01)} & \textbf{96.16(3.15)} \\ \hline
			\multirow{4}{*}{\rotatebox{90}{Target}}     & DaST~\cite{DaST}    & 18.23(1.24)          & 62.23(1.54)          & 41.60(1.17)          & 14.84(5.23)          & 25.07(5.12)          & 20.04(5.12)          & 22.08(4.66)          & 60.80(3.82)          & 45.33(4.05)          \\
			& DST~\cite{DST}      & 23.83(1.50)          & 65.75(1.43)          & 42.15(1.15)          & \textbf{18.43(5.21)} & 24.57(5.22)          & 22.54(4.98)          & 26.97(4.47)          & 70.00(3.92)          & 56.00(3.99)          \\
			& MAZE-PD~\cite{MAZE} & 29.35(1.50)          & 69.20(1.42)          & 41.99(1.08)          & 18.37(5.23)          & 25.02(5.01)          & 24.65(5.10)          & 43.35(5.10)          & 69.57(3.78)          & 56.39(4.05)          \\
			& LCA (Ours)                          & \textbf{31.77(1.55)} & \textbf{74.79(1.42)} & \textbf{48.36(0.99)} & 5.85(4.16)           & \textbf{50.46(3.17)} & \textbf{51.63(3.19)} & \textbf{44.14(4.62)} & \textbf{79.24(3.79)} & \textbf{79.93(3.97)} \\ \hline
		\end{tabular}
	}
\end{table*}
\subsubsection{ASR on different white-box adversarial sample generation methods} 
We present the ASR using different white-box adversarial sample generation methods for both target and non-target attacks in Table~\ref{table2}.
It is evident that our LCA improves in each white-box adversarial sample generation method compared to other schemes. Moreover, higher ASR is achieved when generating adversarial samples using BIM and PGD schemes compared to FGSM due to the better quality of the adversarial samples generated by the former.
Notably, our LCA achieves a relatively high ASR boost in non-target attacks compared to target attacks. This is due to the fact that the data generated by LCA has high accuracy for the target model, which makes the trained substitute model more similar to the target model, and thus the non-target attack is successful. However, the generated data still has negative samples for the target model, and the trained substitute model still has a gap compared to the target model, thus the enhancement in the target attack is not significant.

\begin{table}[!t]
	\centering
	\caption{Comparison of the attack success rates of our approach and competitors in terms of label-only and probability-based scenarios. Using BIM as the default attack method and ResNet18 trained on CIFAR10 as the default target model. The best result is \textbf{bolded}. \label{table3}}
	\resizebox{\linewidth}{!}{
		\begin{tabular}{c|c|cc|cc}
			\hline
			&                                  & \multicolumn{2}{c|}{Probability-based}       & \multicolumn{2}{c}{Label-only} \\
			& \multirow{-2}{*}{Method}         & ASR                          & distance      & ASR            & distance      \\ \hline
			& DaST~\cite{DaST} & 60.45                        & 0.99          & 44.72          & 1.19          \\
			& MAZE~\cite{MAZE} & 78.54                        & 0.98          & -              & -             \\
			& DST~\cite{DST}   & 90.97                        & 0.98          & 78.24          & 1.18          \\
			\multirow{-4}{*}{Non-Target} & LCA (Ours)                        & \textbf{93.33}               & \textbf{0.94} & \textbf{90.65} & \textbf{0.94} \\ \hline
			& DaST~\cite{DaST} & 62.23                        & 1.54          & 46.96          & 1.54          \\
			& MAZE~\cite{MAZE} & 65.75                        & 1.43          & -              & -             \\
			& DST~\cite{DST}   & 69.20                        & 1.42          & 58.04          & 1.54          \\
			\multirow{-4}{*}{Target}     & LCA (Ours)                        & \textbf{74.79} & \textbf{1.42}          & \textbf{70.09}          & \textbf{1.39}          \\ \hline
	\end{tabular}}
\end{table}
\subsubsection{ASR on label-only and probability-based cases} 
We report the ASR for both label-only output and probability-based output of the target model, as shown in Table~\ref{table3}.
Our LCA outperforms existing substitute attack schemes in terms of ASR in both label-only and probability-based output cases. Moreover, our LCA demonstrates similar ASR in both label-only and probability-based cases. This is due to the fact that our LCA does not rely on probabilistic outputs for discriminating member data, such that the distribution of the generated data in both cases closely aligns with the distribution of the training data of the target model. %As a result, the target network exhibits relatively high accuracy on these data samples. 
Even when the target model has label-only outputs, the substitute model can still learn the knowledge of the target model during the substitute training process. As a result, the trained substitute model is closer to the target model in both output cases, leading to outstanding ASR performance.

\subsubsection{Substitute Training Efficiency} 
To investigate the substitute training efficiency, we analyze the number of queries in two stages. In stage 1, for each data generated by the SD, we need to query the target model twice: once for clean images and once for noisy images. The number of queries in stage 1, denoted as $N_1$, depends on the number of classes and the size of the codebook. In stage 2, we train the substitute model by continuously querying the output of the target model, denoted as $N_2$, which depends on factors such as the number of iterations and batch size in the specific training process.

Therefore, the total query budget, denoted as $N_{QB}$, is the sum of the queries from both stages. It can be expressed as:
\begin{equation}
	N_{QB} = N_1 + N_2.
\end{equation}

\begin{table}[!t]
	\centering
	\caption{The query volume of the first stage of our LCA. Experiments are conducted under CIFAR10, CIFAR100 and STL10 datasets, different codebook sizes as well as probability-based and label-only cases, respectively. \label{table4}}
	\resizebox{\linewidth}{!}{
		\begin{tabular}{c|c|ccccc}
			\hline
			& \multirow{2}{*}{Dataset} & \multicolumn{5}{c}{Codebook size}  \\
			&                          & 2    & 5    & 10   & 20    & 50    \\ \hline
			\multirow{3}{*}{Probability-based} & CIFAR10                  & 90   & 206  & 389  & 692   & 1580  \\
			& CIFAR100                 & 1045 & 2548 & 5100 & 10017 & 23900 \\
			& STL10                    & 91   & 194  & 351  & 644   & 1523  \\ \hline
			\multirow{3}{*}{Label-only}        & CIFAR10                  & 63   & 146  & 269  & 525   & 1231  \\
			& CIFAR100                 & 954  & 2241 & 4521 & 8847  & 18560 \\
			& STL10                    & 75   & 169  & 319  & 598   & 1438  \\ \hline
		\end{tabular}
	}
\end{table}
The first stage query number $N_1$ for different datasets at different codebook sizes is provided in Table~\ref{table4}. It is noticed that $N_1$ primarily depends on the number of classes and the size of the codebook. That is, a larger codebook size and more classes will result in a higher number of queries. The impact of codebook size on the ASR will be discussed in Sec. \ref{sec_Ab}.
%Additionally, the precision of the target model can also influence the number of queries required to obtain the full codebook. If the target model has a relatively low accuracy, more queries may be needed to cover all the classes in the codebook. But in this paper, it is assumed that the target model has a relatively high accuracy, and therefore do not consider this factor.

\begin{figure}[!t]
	\centering
	\includegraphics[width=\linewidth]{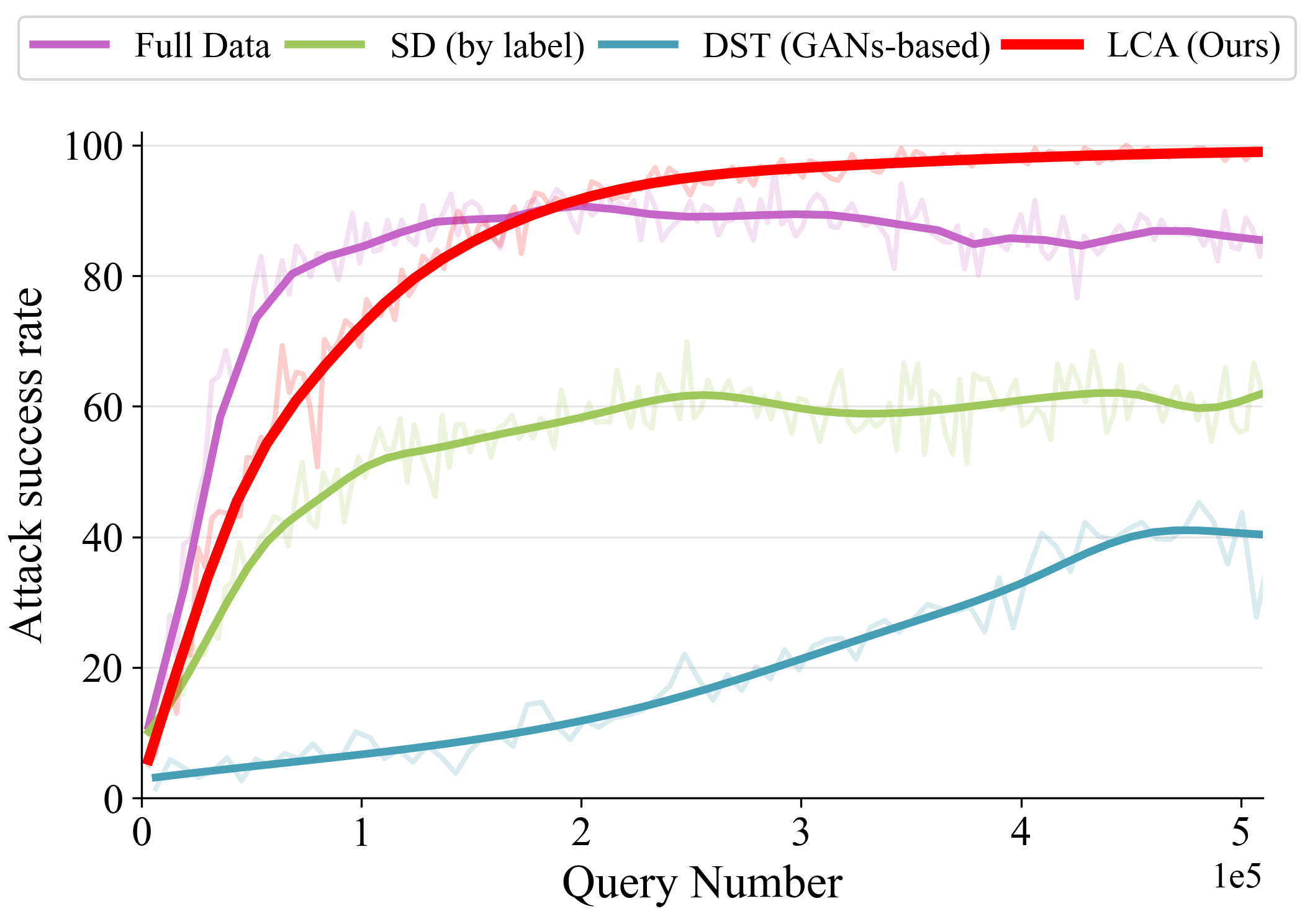}
	\caption{The attack success rates under different scenarios or schemes. The experiments use ResNet18 trained in CIFAR10 as the target model and use BIM to generate adversarial samples.}
	\label{fig_6}
\end{figure}
We discuss the ASR under different query budgets, and the results are given in Fig.~\ref{fig_6} (for ResNet18 trained on the CIFAR10 dataset) and in Fig.~\ref{fig_1} (c) (for ResNet50 trained on the CIFAR100 dataset). The codebook size factor is fixed at 10 in these experiments. We give the results of substitute training using data from different scenarios, which include: data generated by the SD using class labels as prompt (generate by label), training data of the target model (Full data), data generated by GANs (DST~\cite{DST}) scheme and our LCA.

It is worth noting that our LCA achieves high ASR with very small query budgets, already reaching 90\% ASR in around 200k queries. In contrast, DST~\cite{DST} scheme that used GANs structures to generate data requires a query budget of more than 9 M to achieve relatively good ASR.
Even more impressively, our LCA outperforms the full-data scenario at larger query budgets in both experiments. This is because the full training dataset remains relatively small for substitute training (both CIFAR-10 and CIFAR-100 contain 50,000 training data), while our LCA continuously generates data throughout the training process. This provides substitute training with a larger quantity and more diverse training data. Therefore, our LCA demonstrates more efficient performance compared to the full-data scenario.

Furthermore, considering the number of queries for different datasets in Table~\ref{table4}, it can be observed that the query number $N_1$ is relatively small compared to the total query budget $N_{QB}$. This indicates that our proposed solution achieves substantial performance gains at a relatively low cost.
Overall, the results demonstrate the effectiveness and efficiency of our LCA and provide superior ASR at lower query budgets compared to existing substitute attack schemes.

\begin{figure}[!t]
	\centering
	\includegraphics[width=\linewidth]{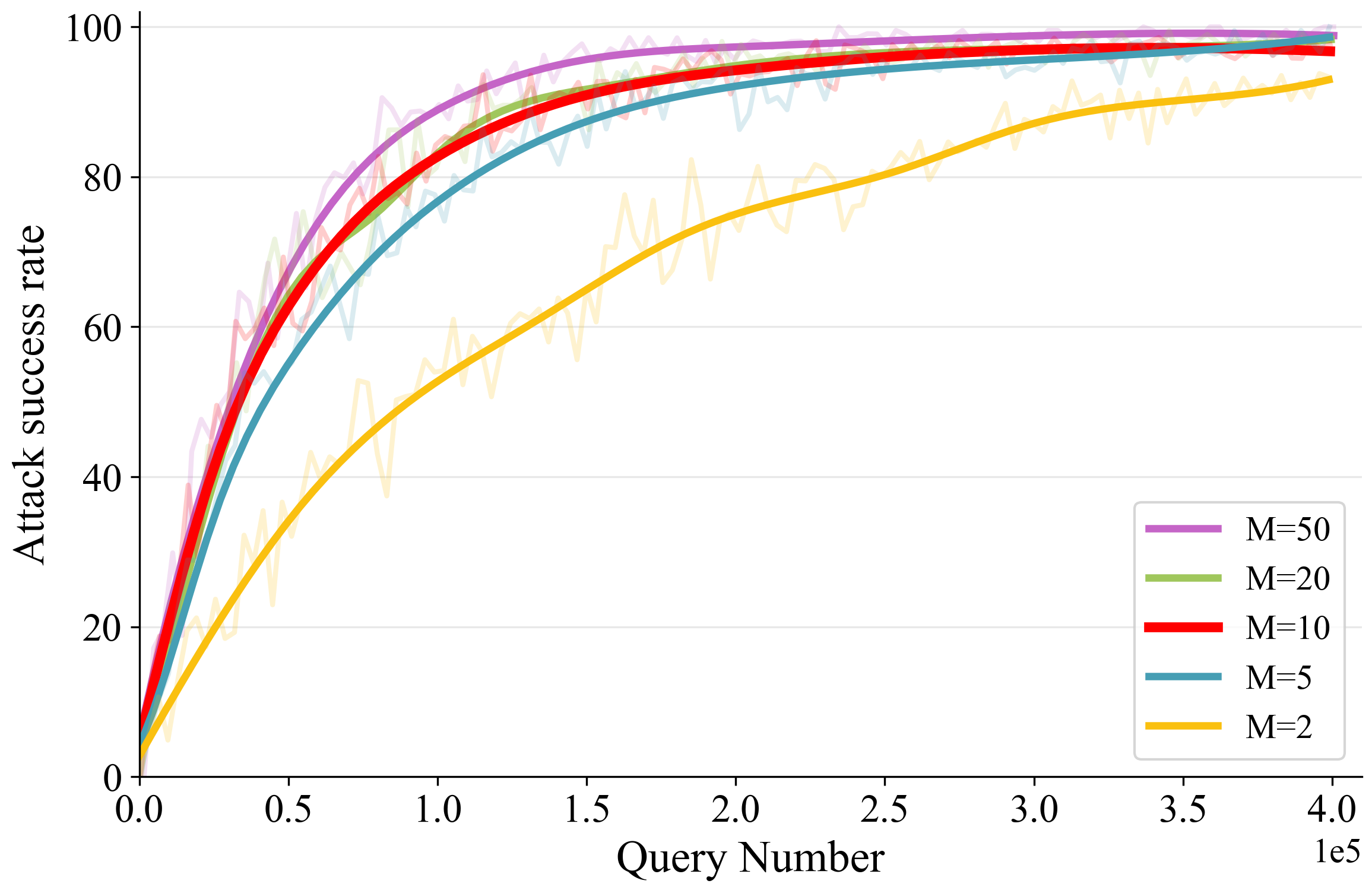}
	\caption{Attack success rates at different codebook sizes. The experiments use ResNet34 trained in STL10 as the target model and use BIM to generate the adversarial samples.}
	\label{fig_7}
\end{figure}
\subsection{Ablation Study}\label{AS}
\label{sec_Ab}
\subsubsection{The efficacy of the codebook size} 
In Fig.~\ref{fig_7}, we provide the effect of different codebook sizes on the ASR. The results show that the codebook size has an impact on the ASR and the number of queries.
When the codebook length is 2, the ASR value of this scheme is lower, and more queries are required. This is because there are too few latent codes available, resulting in insufficient diversity of the generated data and thus lower training efficiency of the network. 
As the codebook length increases to 5, the ASR improves for the same number of queries. Similarly, the scheme becomes more effective when the codebook length is further increased to 10.
However, when the codebook length is set to 20 and 50, the increase in ASR is minimal. This is due to the fact that a codebook size of 10 already generates sufficiently diverse data for substitute training, and a larger codebook is merely redundant without playing a key role.
Combining with Table~\ref{table4}, more query budget is required when the codebook length is larger.
Therefore, considering both the ASR and the query budget, we believe that setting the codebook size to 10 in our LCA is optimal. With this setting, LCA generates sufficiently diverse images and achieves good performance in terms of ASR and query consumption.

\subsubsection{The efficacy of different components}
The ablation experiments examining the effect of different components on the ASR are presented in Table~\ref{table5}. The experiments are conducted using ResNet50 trained on the CIFAR100 dataset, and the default white-box adversarial sample generation method is BIM.
In Table~\ref{table5}, the `Baseline' refers to the scheme that uses labels as the prompt for the SD to generate data for substitute training. The `w/o LCA' indicates that LCA is not used and training is performed with inferred member data. 
\begin{table}[!t]
	\centering
	\caption{The attack success rates of models with different settings. The experiments use ResNet50 trained in CIFAR100 as the target model and use BIM to generate the adversarial samples.`Baseline' means that the data is generated directly using the SD for training the substitute model, `w/o LCA' indicates that LCA is not used and training is performed with inferred member data. The best result is \textbf{bolded}. \label{table5}}
	\resizebox{\linewidth}{!}{
		\begin{tabular}{c|c|cc|cc}
			\hline
			& \multirow{2}{*}{Method} & \multicolumn{2}{c|}{Probability-based} & \multicolumn{2}{c}{Label-only} \\
			&                         & ASR                & distance          & ASR             & distance      \\ \hline
			\multirow{3}{*}{Non-Target} & Baseline                & 60.85              & 4.05              & 61.88           & 4.06          \\
			& w/o LCA              & 65.51              & 4.07              & 66.34           & 4.06          \\
			& LCA (Ours)              & \textbf{97.59}     & \textbf{4.06}     & \textbf{97.98}  & \textbf{4.06} \\ \hline
			\multirow{3}{*}{Target}     & Baseline                & 21.47              & 3.15              & 15.48           & 3.19          \\
			& w/o LCA              & 25.24              & 3.15              & 18.78           & 3.19          \\
			& LCA (Ours)              & \textbf{50.46}     & \textbf{3.17}     & \textbf{47.46}  & \textbf{3.18} \\ \hline
	\end{tabular}}
\end{table}

The results demonstrate that the data generated directly from the SD is less accurate for the specific target model, resulting in relatively low ASR. In contrast, with the use of our LCA, the ASR is significantly improved.
Overall, these ablation experiments highlight the importance of latent code augmentation in improving the ASR.

\section{Conclusion}
In this paper, we propose a novel data-free substitute attack scheme based on the SD to improve the efficiency and accuracy of substitute training. Compared to existing GANs-based schemes that generate images with low quality and require re-training for each target model, we alleviate these problems by utilizing the pre-trained Stable Diffusion to generate substitute training data in data-free scenarios.
In addition, to further facilitate the SD to generate data that conforms to the data distribution of the target model, we propose a novel LCA to guide the SD for generating data.
Subsequently, the generated data is used to train the substitute model, which allows the substitute model to simulate the target model more efficiently.
Benefiting from the guidance of LCA, SD generates diverse data conforming to the data distribution for the target model. This effectively improves the efficiency of the substitute training and ensures that the substitute training can be applied to a variety of target models trained on different datasets.
Numerous experiments demonstrate that our approach achieves a higher attack success rate while reducing the number of queries compared to existing substitute attack schemes.

\section*{Acknowledgments}
This work was supported in part by the National Key Research and development Program of China (2021YFA1000102), and in part by the grants from the National Natural Science Foundation of China (Nos. 62376285, 61673396), Natural Science Foundation of Shandong Province (No: ZR2022MF260).

\section*{Ethics Statement}
The data-free substitute attacks presented in this paper focus on exploring neural network security and robustness. We strictly adhere to ethical reviews and laws and regulations, reduce the risk of potential misuse, and increase transparency and fairness. We actively promote the legitimacy and fairness of research activities to minimize potential risks and harm.

\bibliographystyle{IEEEtran}
\bibliography{Attack23.bib}

% Generated by IEEEtran.bst, version: 1.14 (2015/08/26)
\begin{thebibliography}{10}
\providecommand{\url}[1]{#1}
\csname url@samestyle\endcsname
\providecommand{\newblock}{\relax}
\providecommand{\bibinfo}[2]{#2}
\providecommand{\BIBentrySTDinterwordspacing}{\spaceskip=0pt\relax}
\providecommand{\BIBentryALTinterwordstretchfactor}{4}
\providecommand{\BIBentryALTinterwordspacing}{\spaceskip=\fontdimen2\font plus
\BIBentryALTinterwordstretchfactor\fontdimen3\font minus
  \fontdimen4\font\relax}
\providecommand{\BIBforeignlanguage}[2]{{%
\expandafter\ifx\csname l@#1\endcsname\relax
\typeout{** WARNING: IEEEtran.bst: No hyphenation pattern has been}%
\typeout{** loaded for the language `#1'. Using the pattern for}%
\typeout{** the default language instead.}%
\else
\language=\csname l@#1\endcsname
\fi
#2}}
\providecommand{\BIBdecl}{\relax}
\BIBdecl

\bibitem{yuan_adversarial_2019}
X.~Yuan, P.~He, Q.~Zhu, and X.~Li, ``Adversarial {Examples}: {Attacks} and
  {Defenses} for {Deep} {Learning},'' \emph{IEEE Transactions on Neural
  Networks and Learning Systems}, vol.~30, no.~9, pp. 2805--2824, Sep. 2019.

\bibitem{MIM}
Y.~Dong, F.~Liao, T.~Pang, H.~Su, J.~Zhu, X.~Hu, and J.~Li, ``Boosting
  adversarial attacks with momentum,'' in \emph{Proceedings of the IEEE/CVF
  Conference on Computer Vision and Pattern Recognition}, 2018, pp. 9185--9193.

\bibitem{UAP}
S.~M. Moosavi~Dezfooli, A.~Fawzi, O.~Fawzi, and P.~Frossard, ``Universal
  adversarial perturbations,'' in \emph{Proceedings of the IEEE/CVF Conference
  on Computer Vision and Pattern Recognition}, 2017, pp. 1765--1773.

\bibitem{katzir_gradients_2021}
Z.~Katzir and Y.~Elovici, ``Gradients {Cannot} {Be} {Tamed}: {Behind} the
  {Impossible} {Paradox} of {Blocking} {Targeted} {Adversarial} {Attacks},''
  \emph{IEEE Transactions on Neural Networks and Learning Systems}, vol.~32,
  no.~1, pp. 128--138, Jan. 2021.

\bibitem{AdverDefen}
K.~Ren, T.~Zheng, Z.~Qin, and X.~Liu, ``Adversarial attacks and defenses in
  deep learning,'' \emph{Engineering}, vol.~6, no.~3, pp. 346--360, 2020.

\bibitem{FGSM}
I.~Goodfellow, J.~Shlens, and C.~Szegedy, ``Explaining and harnessing
  adversarial examples,'' in \emph{International Conference on Machine
  Learning}, 2015.

\bibitem{BIM}
A.~Kurakin, I.~Goodfellow, and S.~Bengio, ``Adversarial examples in the
  physical world,'' in \emph{International Conference on Learning
  Representations}, 2017.

\bibitem{PGD}
A.~Madry, A.~Makelov, L.~Schmidt, D.~Tsipras, and A.~Vladu, ``Towards deep
  learning models resistant to adversarial attacks,'' in \emph{International
  Conference on Learning Representations}, 2018.

\bibitem{yue2021black}
Z.~Yue, Z.~He, H.~Zeng, and J.~McAuley, ``Black-box attacks on sequential
  recommenders via data-free model extraction,'' in \emph{Proceedings of the
  15th ACM Conference on Recommender Systems}, 2021, pp. 44--54.

\bibitem{EBA}
Z.~Cai, Y.~Tan, and M.~S. Asif, ``Ensemble-based blackbox attacks on dense
  prediction,'' in \emph{Proceedings of the IEEE/CVF Conference on Computer
  Vision and Pattern Recognition}, 2023.

\bibitem{TSEA}
H.~Huang, Z.~Chen, H.~Chen, Y.~Wang, and K.~Zhang, ``{T-SEA}: Transfer-based
  self-ensemble attack on object detection,'' in \emph{Proceedings of the
  IEEE/CVF Conference on Computer Vision and Pattern Recognition}, 2023.

\bibitem{MAZE}
S.~Kariyappa, A.~Prakash, and M.~K. Qureshi, ``{MAZE}: Data-free model stealing
  attack using zeroth-order gradient estimation,'' in \emph{Proceedings of the
  IEEE/CVF Conference on Computer Vision and Pattern Recognition}, 2021, pp.
  13\,814--13\,823.

\bibitem{DaST}
M.~Zhou, J.~Wu, Y.~Liu, S.~Liu, and C.~Zhu, ``{DaST}: Data-free substitute
  training for adversarial attacks,'' in \emph{Proceedings of the IEEE/CVF
  Conference on Computer Vision and Pattern Recognition}, 2020, pp. 234--243.

\bibitem{DST}
W.~Wang, X.~Qian, Y.~Fu, and X.~Xue, ``{DST}: Dynamic substitute training for
  data-free black-box attack,'' in \emph{Proceedings of the IEEE/CVF Conference
  on Computer Vision and Pattern Recognition}, 2022, pp. 14\,361--14\,370.

\bibitem{HardLabel}
S.~Sanyal, S.~Addepalli, and R.~V. Babu, ``Towards data-free model stealing in
  a hard label setting,'' in \emph{Proceedings of the IEEE/CVF Conference on
  Computer Vision and Pattern Recognition}, 2022, pp. 15\,284--15\,293.

\bibitem{ColorQuery}
S.~Yuan, Q.~Zhang, L.~Gao, Y.~Cheng, and J.~Song, ``Natural color fool: Towards
  boosting black-box unrestricted attacks,'' \emph{Advances in Neural
  Information Processing Systems}, vol.~35, pp. 7546--7560, 2022.

\bibitem{queries}
Y.~Senzaki, S.~Ohata, and K.~Matsuura, ``Simple black-box adversarial examples
  generation with very few queries,'' \emph{IEICE Transactions on Information
  and Systems}, vol. 103, no.~2, pp. 212--221, 2020.

\bibitem{9727149}
X.~Li, X.~Zhang, F.~Yin, and C.~Liu, ``Decision-based adversarial attack with
  frequency mixup,'' \emph{IEEE Transactions on Information Forensics and
  Security}, vol.~17, pp. 1038--1052, 2022.

\bibitem{TED}
J.~Zhang, B.~Li, J.~Xu, S.~Wu, S.~Ding, L.~Zhang, and C.~Wu, ``Towards
  efficient data free black-box adversarial attack,'' in \emph{Proceedings of
  the IEEE/CVF Conference on Computer Vision and Pattern Recognition}, 2022,
  pp. 15\,094--15\,104.

\bibitem{DDG}
W.~Wang, B.~Yin, T.~Yao, L.~Zhang, Y.~Fu, S.~Ding, J.~Li, F.~Huang, and X.~Xue,
  ``Delving into data: Effectively substitute training for black-box attack,''
  in \emph{Proceedings of the IEEE/CVF Conference on Computer Vision and
  Pattern Recognition}, 2021, pp. 4761--4770.

\bibitem{DDPM}
J.~Ho, A.~Jain, and P.~Abbeel, ``Denoising diffusion probabilistic models,''
  \emph{arXiv preprint arXiv:2006.11239}, 2020.

\bibitem{StableDiffusion}
R.~Rombach, A.~Blattmann, D.~Lorenz, P.~Esser, and B.~Ommer, ``High-resolution
  image synthesis with latent diffusion models,'' in \emph{Proceedings of the
  IEEE/CVF Conference on Computer Vision and Pattern Recognition}, 2022.

\bibitem{DistrWhite}
Y.~Zhu, Y.~Chen, X.~Li, K.~Chen, Y.~He, X.~Tian, B.~Zheng, Y.~Chen, and
  Q.~Huang, ``Toward understanding and boosting adversarial transferability
  from a distribution perspective,'' \emph{IEEE Transactions on Image
  Processing}, vol.~31, pp. 6487--6501, 2022.

\bibitem{chen2017zoo}
P.~Chen, H.~Zhang, Y.~Sharma, J.~Yi, and C.~Hsieh, ``Zoo: Zeroth order
  optimization based black-box attacks to deep neural networks without training
  substitute models,'' in \emph{Proceedings of the 10th ACM workshop on
  Artificial Intelligence and Security}, 2017, pp. 15--26.

\bibitem{hu2022substitute}
C.~Hu, H.~Xu, and X.~Wu, ``Substitute meta-learning for black-box adversarial
  attack,'' \emph{IEEE Signal Processing Letters}, vol.~29, pp. 2472--2476,
  2022.

\bibitem{zhou2022adversarial}
L.~Zhou, P.~Cui, X.~Zhang, Y.~Jiang, and S.~Yang, ``Adversarial eigen attack on
  black-box models,'' in \emph{Proceedings of the IEEE/CVF Conference on
  Computer Vision and Pattern Recognition}, 2022, pp. 15\,254--15\,262.

\bibitem{duan_dual_2023}
M.~Duan, Y.~Qin, J.~Deng, K.~Li, and B.~Xiao, ``Dual {Attention} {Adversarial}
  {Attacks} {With} {Limited} {Perturbations},'' \emph{IEEE Transactions on
  Neural Networks and Learning Systems}, pp. 1--15, 2023.

\bibitem{Sibling}
Z.~Li, B.~Yin, T.~Yao, J.~Guo, S.~Ding, S.~Chen, and C.~Liu, ``Sibling-attack:
  Rethinking transferable adversarial attacks against face recognition,'' in
  \emph{Proceedings of the IEEE/CVF Conference on Computer Vision and Pattern
  Recognition}, 2023.

\bibitem{cui2020substitute}
W.~Cui, X.~Li, J.~Huang, W.~Wang, S.~Wang, and J.~Chen, ``Substitute model
  generation for black-box adversarial attack based on knowledge
  distillation,'' in \emph{2020 IEEE International Conference on Image
  Processing}, 2020, pp. 648--652.

\bibitem{park2020partial}
H.~Park, G.~Ryu, and D.~Choi, ``Partial retraining substitute model for
  query-limited black-box attacks,'' \emph{Applied Sciences}, vol.~10, no.~20,
  p. 7168, 2020.

\bibitem{zhu2022defense}
Z.~Zhu, B.~Zhu, H.~Zhang, Y.~Geng, L.~Wang, D.~Zhang, and Z.~Gu, ``Defense
  against query-based black-box attack with small gaussian-noise,'' in
  \emph{2022 7th IEEE International Conference on Data Science in Cyberspace},
  2022, pp. 249--256.

\bibitem{LOGAN}
G.~D. E. D.~C. Jamie~Hayes, Luca~Melis, ``{LOGAN}: Membership inference attacks
  against generative models,'' \emph{Proceedings on Privacy Enhancing
  Technologies}, vol.~1, pp. 133--152, 2019.

\bibitem{GANLeaks}
D.~Chen, N.~Yu, Y.~Zhang, and M.~Fritz, ``{GAN-Leaks}: A taxonomy of membership
  inference attacks against generative models,'' in \emph{Proceedings of the
  2020 ACM SIGSAC Conference on Computer and Communications Security}.\hskip
  1em plus 0.5em minus 0.4em\relax ACM, 2020.

\bibitem{MLLeaks}
A.~Salem, Y.~Zhang, M.~Humbert, P.~Berrang, M.~Fritz, and M.~Backes,
  ``{ML-Leaks}: Model and data independent membership inference attacks and
  defenses on machine learning models,'' in \emph{Network and Distributed
  System Security Symposium}, 2019.

\bibitem{LabelonlyMI}
C.~A. Choquette-Choo, F.~Tram`er, N.~Carlini, and N.~Papernot, ``Label-only
  membership inference attacks,'' in \emph{Proceedings of the 38th
  International Conference on Machine Learning}, 2021.

\bibitem{LAION-5B}
C.~Schuhmann, R.~Beaumont, R.~Vencu, C.~Gordon, R.~Wightman, M.~Cherti,
  T.~Coombes, A.~Katta, C.~Mullis, M.~Wortsman, P.~Schramowski, S.~Kundurthy,
  K.~Crowson, L.~Schmidt, R.~Kaczmarczyk, and J.~Jitsev, ``{LAION}-{5B}: {An}
  open large-scale dataset for training next generation image-text models,''
  \emph{Advances in Neural Information Processing Systems}, vol.~35, pp.
  25\,278--25\,294, 2022.

\bibitem{CLIP}
A.~Radford, J.~W. Kim, C.~Hallacy, A.~Ramesh, G.~Goh, S.~Agarwal, G.~Sastry,
  A.~Askell, P.~Mishkin, J.~Clark, G.~Krueger, and I.~Sutskever, ``Learning
  transferable visual models from natural language supervision,'' in
  \emph{International Conference on Machine Learning}, vol. 139, 2021, pp.
  8748--8763.

\bibitem{MixUp}
H.~Zhang, M.~Cisse, Y.~N. Dauphin, and D.~Lopez-Paz, ``{MixUp}: {Beyond}
  empirical risk minimization,'' \emph{arXiv preprint arxiv:1710.09412}, 2017.

\bibitem{CutMix}
S.~Yun, D.~Han, S.~Chun, S.~J. Oh, Y.~Yoo, and J.~Choe, ``{CutMix}:
  Regularization strategy to train strong classifiers with localizable
  features,'' in \emph{2019 {IEEE}/{CVF} {International} {Conference} on
  {Computer} {Vision} ({ICCV})}.\hskip 1em plus 0.5em minus 0.4em\relax IEEE,
  2019, pp. 6022--6031.

\bibitem{RICAP}
R.~Takahashi, T.~Matsubara, and K.~Uehara, ``Data augmentation using random
  image cropping and patching for deep cnns,'' \emph{IEEE Transactions on
  Circuits and Systems for Video Technology}, vol.~30, no.~9, pp. 2917--2931,
  Sep. 2020.

\bibitem{equiv4}
D.~E. Worrall, S.~J. Garbin, and D.~T. G.~J. Brostow, ``Harmonic networks: Deep
  translation and rotation equivariance,'' in \emph{Proceedings of the IEEE/CVF
  Conference on Computer Vision and Pattern Recognition}, 2017.

\bibitem{equiv}
N.~Khetan, T.~Arora, S.~U. Rehman, and D.~K. Gupta, ``Implicit equivariance in
  convolutional networks,'' \emph{arXiv preprint arXiv:2111.14157}, 2021.

\bibitem{equiv1}
A.~Nasiri and T.~Bepler, ``Unsupervised object representation learning using
  translation and rotation group equivariant vae,'' in \emph{36th Conference on
  Neural Information Processing Systems}, 2022.

\bibitem{ResNet}
K.~He, X.~Zhang, S.~Ren, and J.~Sun, ``Deep residual learning for image
  recognition,'' in \emph{Proceedings of the IEEE/CVF Conference on Computer
  Vision and Pattern Recognition}, 2016, pp. 770--778.

\bibitem{VGG}
K.~Simonyan and A.~Zisserman, ``Very deep convolutional networks for
  large-scale image recognition,'' \emph{arXiv preprint arXiv:1409.1556}, 2014.

\bibitem{CIFAR}
A.~Krizhevsky, ``Learning multiple layers of features from tiny images,''
  \emph{Master's thesis, University of Tront}, 2009.

\bibitem{STL10}
A.~Coates, H.~Lee, and A.~Y. Ng, ``An analysis of single layer networks in
  unsupervised feature learning,'' in \emph{Proceedings of the Fourteenth
  International Conference on Artificial Intelligence and Statistics.}, 2011.

\bibitem{Tinyimagenet}
O.~Russakovsky, J.~Deng, H.~Su, J.~Krause, S.~Satheesh, S.~Ma, Z.~Huang,
  A.~Karpathy, A.~Khosla, M.~Bernstein, A.~C. Berg, and L.~Fei-Fei, ``Imagenet
  large scale visual recognition challenge,'' \emph{International Journal of
  Computer Vision}, vol. 115, no.~3, pp. 211--252, 2015.

\end{thebibliography}

\vfill

\end{document}